\documentclass[runningheads]{llncs}

\usepackage{makecell}
\usepackage{graphicx}
\usepackage{multirow}
\usepackage{amsmath,amssymb,amsfonts}
\usepackage{mathrsfs}
\usepackage[title]{appendix}
\usepackage{xcolor}
\usepackage{textcomp}
\usepackage{manyfoot}
\usepackage{booktabs}
\usepackage{listings}
\usepackage[most]{tcolorbox}
\usepackage{adjustbox,threeparttable}
\usepackage[utf8]{inputenc}
\usepackage{placeins}
\usepackage{float}
\usepackage{url}
\usepackage{cite}

\usepackage{hyperref}

\begin{document}

\title{SurgCheck: Do Vision--Language Models Really Look at Images in Surgical VQA?}

\titlerunning{SurgCheck}

\author{Jongmin Shin$^{\dagger}$\inst{1}
\and Ka Young Kim$^{\dagger}$\inst{2}
\and Eunki Cho\inst{2}
\and Seong Tae Kim$^{*}$\inst{2}
\and Namkee Oh$^{*}$\inst{1}}

\authorrunning{J. Shin et al.}

\institute{Department of Surgery, Samsung Medical Center, Seoul 06351, Republic of Korea
\and
Kyung Hee University, Yongin 17104, Republic of Korea}

\maketitle
\renewcommand{\thefootnote}{}\footnote{$^{\dagger}$ Equal contribution; $^{*}$ Corresponding author.}\setcounter{footnote}{0}\renewcommand{\thefootnote}{\arabic{footnote}}

\begin{abstract}
\textbf{Purpose:} Vision--language models (VLMs) have shown promising performance in surgical 
visual question answering (VQA). However, existing surgical VQA datasets often 
contain linguistic shortcuts, where question phrasing implicitly constrains 
the answer space. In safety-critical surgical settings, it remains unclear 
whether reported performance reflects visual understanding or reliance on such 
linguistic shortcuts. \textbf{Methods:} We introduce SurgCheck, a diagnostic benchmark for quantifying 
linguistic shortcut reliance in surgical VQA. SurgCheck employs a paired-question 
design in which each surgical frame is associated with an original question 
containing entity names and a less-biased counterpart that removes these 
names while preserving identical visual content and ground-truth answers. 
The resulting performance gap provides a diagnostic signal of shortcut 
reliance. To ensure that the less-biased question remains well-defined even without entity names, four 
grounding cues are incorporated: bounding box, arrow, spatial position, and 
periphrasis. We evaluate both general-purpose and surgical-specific VLMs under zero-shot and fine-tuned settings on SurgCheck. To evaluate open-ended zero-shot responses, we introduce an LLM-as-a-judge evaluation protocol. \textbf{Results:} Using SurgCheck, we observe consistent performance degradation on less-biased questions across five VLMs, despite identical visual inputs. Text-only ablation reveals minimal performance drops for action and target prediction, indicating that action and target prediction is largely driven by linguistic shortcuts rather than visual reasoning. \textbf{Conclusion:} SurgCheck provides a controlled diagnostic framework that exposes failure modes masked by linguistic bias in existing surgical VQA benchmarks. Our findings demonstrate that strong benchmark performance does not necessarily imply faithful visual understanding, underscoring the need for bias-aware evaluation in surgical VQA. SurgCheck is publicly available at \href{https://github.com/ailab-kyunghee/SurgCheck}{https://github.com/ailab-kyunghee/SurgCheck}.

\end{abstract}

\section{Introduction}\label{intro}
Recent advances in vision–language models (VLMs) have led to strong reported performance in surgical visual question answering (VQA), suggesting potential value for intraoperative decision support and surgical education \cite{cholec80vqa,endochat,ssg_vqa,surgical_vqla,med_vqa_survey,psi_ava_vqa,he2024pitvqa,surgen_net}. However, it remains unclear whether such performance reflects genuine visual reasoning or exploitation of linguistic shortcuts embedded in question phrasing. Many existing surgical VQA datasets \cite{cholec80vqa,ssg_vqa,endochat} include questions that explicitly mention answer-correlated entities (e.g., “What action is the bipolar performing?”), enabling models to infer answers from linguistic shortcuts rather than visual evidence. As a result, benchmark accuracy may not reliably indicate faithful visual understanding.
Recent studies have attempted to mitigate linguistic bias in surgical VQA through improved dataset construction and model design. On the dataset side, prior work expanded question diversity and introduced bias-reduction heuristics such as template filtering and scene-graph–based question generation \cite{ssg_vqa}, while model-side approaches incorporated enhanced visual encoders and contrastive reasoning to improve grounding \cite{endochat}. However, the challenge persists in surgical settings, where multiple similar instruments often interact with anatomically proximal regions. As a result, explicit entity references remain useful for human clarity but potentially exploitable by models, leaving it unclear how much reported performance reflects genuine visual reasoning. Related issues have also been studied in general-domain VQA, where benchmarks such as VQA-CP \cite{agrawal2018don} show that models exploit question priors rather than visual evidence. Unlike prior work that focuses on mitigating such biases during training, our work focuses on diagnosing linguistic shortcut reliance at evaluation time, particularly in the surgical domain where instrument–action conventions induce domain-specific shortcut patterns.

\begin{figure}[t]
\centering
\includegraphics[width=\textwidth]{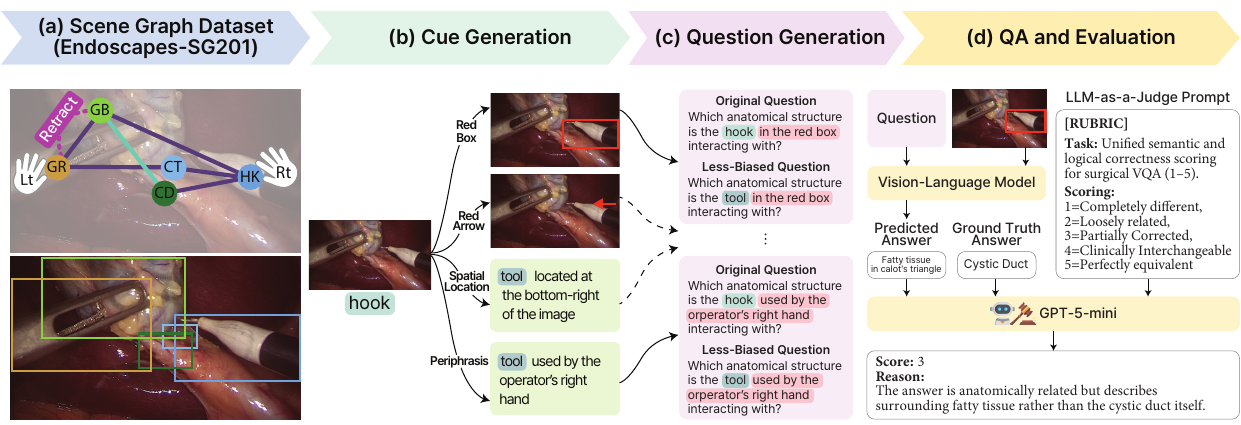}
\caption{\textbf{Overview of the SurgCheck dataset and evaluation pipeline.} The dataset construction begins with structured scene-graph annotations from Endoscapes-SG201, followed by paired-question generation for each grounding cue (e.g., red box or arrow). Each surgical frame is paired with an \textit{original} and a \textit{less-biased} question variant, differing only in linguistic bias. During evaluation, a vision–language model (VLM) predicts answers that are semantically judged by a large language model (LLM) using a rubric-based scoring protocol.}
\label{fig:overall_pipeline}
\end{figure}

To address this evaluation gap, we introduce \textbf{SurgCheck}, a diagnostic benchmark designed to quantify linguistic shortcut reliance in surgical VQA through a bias-controlled paired-question design. Each surgical frame is associated with two semantically equivalent questions: an \textit{original} version that follows conventional surgical phrasing with explicit entity names, and a \textit{less-biased} counterpart that removes these textual hints while preserving identical visual content and ground-truth answers. Because both question variants share the same image and answer and differ only in the presence of linguistic shortcuts, the resulting performance gap provides a diagnostic signal of whether a model relies on language cues or genuine visual evidence.


To ensure that less-biased questions remain well-defined after removing explicit entity names, SurgCheck leverages four cue types: bounding box, arrow, spatial position, and periphrasis. Each cue provides referential grounding for the queried object, either visually or descriptively. By controlling the type of grounding cue while keeping the visual content and answer fixed, SurgCheck enables a principled analysis of how grounding explicitness affects reliance on visual evidence.

We evaluated five models, three general-purpose VLMs (Qwen2.5-VL~\cite{qwen2_5},
LLaVA-OneVision~\cite{llava1_5}, and PaliGemma2~\cite{paligemma2}) and two
surgical-specific models (PitVQA-Net~\cite{he2024pitvqa} and
EndoChat~\cite{endochat}) under both zero-shot and fine-tuned settings.
SurgCheck exposes systematic differences between original and
less-biased questions under identical visual inputs, supporting its
use as a diagnostic framework for linguistic shortcut reliance in surgical
VQA. To evaluate open-ended zeroshot outputs, we introduce an \textit{LLM-as-a-Judge} protocol for semantic assessment as detailed in Section~\ref{sec:eval_protocol}.
We summarize our main contributions as follows:
\begin{itemize}
    \item \textbf{Bias-Controlled Paired-Question Benchmark for Surgical VQA.}
    We introduce \textit{SurgCheck}, a novel diagnostic benchmark featuring
    a bias-controlled paired-question design with grounding cues,
    where original and less-biased questions share identical visual inputs
    and answers.
    
    \item \textbf{Quantitative Analysis of Performance Gaps on SurgCheck.}
    We evaluate five vision–language models across cue types and
question categories, measuring shortcut reliance via performance
gaps between original and less-biased questions.

    \item \textbf{Linguistic Shortcut Diagnosis via Text-Only Ablation.}
We conduct a text-only ablation experiment, in which visual input is removed, to analyze how VLMs utilize visual information across different question categories.
Our analysis shows that, for original questions, performance remains high even without visual input, particularly in the action category, suggesting that models often rely more on question text than on visual evidence.
\end{itemize}
\section{Methods}\label{methods}
An overview of the SurgCheck diagnostic data construction and evaluation process is illustrated in Fig.~\ref{fig:overall_pipeline}. The pipeline comprises four main stages: (i) dataset construction based on scene-graph annotations from Endoscapes-SG201~\cite{ssg201}, (ii) grounding cue generation to associate each question with its region of interest, (iii) paired-question generation producing original and less-biased variants, and (iv) evaluation using an LLM-as-a-Judge framework.


\subsection{SurgCheck Dataset} 
We construct \textbf{SurgCheck}, a bias-controlled benchmark designed to evaluate linguistic shortcut reliance and visual understanding in surgical visual question answering (VQA). SurgCheck is built on top of the publicly available \textbf{Endoscapes-SG201} dataset, which provides frame-level scene-graph annotations and Critical View of Safety (CVS) labels for laparoscopic cholecystectomy videos. As illustrated in Fig.~\ref{fig:overall_pipeline}, each frame contains a ground-truth scene graph describing surgical instruments and anatomical structures (\textit{nodes}) and their relationships (\textit{edges}) within the operative field. Nodes are annotated with bounding boxes and attributes such as the operating hand, while edges encode spatial and action relations. These structured annotations enable systematic generation of diverse question--answer (QA) pairs while allowing precise control over visual content and semantic intent. SurgCheck contains 1,933 frames and 29,635 original QA pairs spanning eight task categories covering observation, composition, and reasoning levels, including Instrument Count, Anatomy, Instrument, Action, Target, Triplet, Anatomy-hop, and CVS. These task categories are used as analytical axes for evaluating model behavior rather than as a proposed task taxonomy. For dataset splitting, we construct SurgCheck training set using the train/validation splits of Endoscapes-SG201, and SurgCheck test set using Endoscapes-SG201 test split.

\subsection{Task Definition} 
Building upon the structured scene-graph annotations, SurgCheck defines eight task types to support systematic analysis of vision–language model (VLM) behavior across different reasoning requirements in surgical VQA. These task categories serve as analytical axes rather than a proposed task taxonomy. The tasks span from low-level observation to high-level procedural reasoning. At the observation level, Instrument Count, Anatomy, Instrument, Action, and Target assess a model’s ability to recognize visual entities, actions, and attributes within the surgical field. The compositional level includes Triplet and Anatomy-hop tasks, which require integrating multiple relational entities. In particular, Anatomy-hop questions involve identifying an anatomical structure relative to another, thereby requiring spatial reasoning beyond direct object recognition. Finally, the reasoning level focuses on the CVS task, which demands multi-step procedural reasoning to determine whether the Critical View of Safety criteria are satisfied. This hierarchical formulation enables analysis of how linguistic shortcuts and visual reasoning affect model performance across different types of surgical understanding.

\subsection{Question Generation} 
\begin{figure}[t]
\centering
\includegraphics[width=\textwidth]{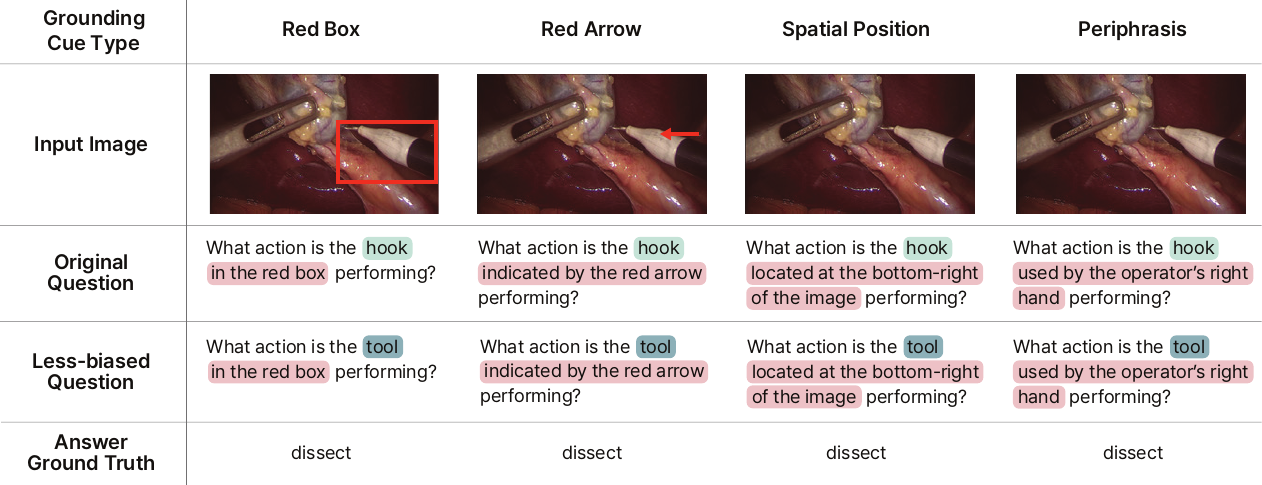}
\caption{\textbf{Four grounding cue types in SurgCheck.} Each cue type demonstrates how the corresponding question pairs differ between the \textit{original} and \textit{less-biased} versions. Red box and red arrow localize the target visually, while spatial position and periphrasis reference it textually through positional or contextual phrasing.}
\label{fig:indirect_cues}
\end{figure}

To systematically evaluate linguistic shortcut reliance in surgical VQA, SurgCheck generates paired question variants for each annotated frame: an original version and a less-biased counterpart. Original questions follow conventional surgical VQA phrasing and explicitly reference entities such as instruments, actions, or anatomical structures. In contrast, less-biased questions preserve the same semantic intent and ground-truth answer while removing explicit entity names, replacing them with neutral referential expressions grounded via grounding cues. Both question variants are automatically generated from the same structured scene-graph annotations and share identical visual inputs and answer space. Consequently, any performance difference between the original and less-biased questions provides a direct diagnostic signal of whether a model relies on linguistic shortcuts rather than genuine visual reasoning. Concrete examples of original and less-biased question pairs across the four grounding cue types are shown in Fig.~\ref{fig:indirect_cues}. This paired-question design is applied to task categories where explicit entity references can introduce linguistic shortcuts, namely Instrument, Action, Target, Triplet, and Anatomy-hop. For Instrument Count, Anatomy, and CVS, questions do not reference specific entities; therefore, original and less-biased variants are identical and serve as shared baselines unaffected by linguistic bias control.

\subsection{Grounding Cue Design} 
In surgical scenes, multiple instruments often appear simultaneously, each performing distinct actions on visually similar anatomical regions. When explicit entity names are removed from a question, it becomes ambiguous which region or object the question refers to. To resolve this ambiguity in a bias-controlled manner, SurgCheck introduces four types of grounding cues that link each question to its corresponding region of interest, as illustrated in Fig.~\ref{fig:indirect_cues}. The cue types span a spectrum of grounding explicitness. Bounding box and arrow provide direct visual localization of the target region, while spatial position describes the target’s approximate location within the image frame (e.g., ``bottom-right of the image''). Periphrasis identifies the target through contextual descriptions, such as its association with the operator’s hands (e.g., ``the tool used by the operator’s right hand''), without revealing its semantic identity. This design ensures that less-biased questions remain unambiguous after removing entity names, while avoiding the introduction of additional semantic hints. Consequently, SurgCheck enables controlled analysis of how grounding explicitness affects model reliance on visual evidence rather than linguistic shortcuts~\cite{vip_llava,r_llava,circle_iccv}.

\subsection{Evaluation Protocol}\label{sec:eval_protocol}
Evaluating vision--language models (VLMs) in surgical VQA is challenging because zero-shot models typically produce free-form natural language responses rather than fixed categorical labels. In this setting, exact-match or rule-based evaluation is unreliable, as semantically correct answers may differ lexically due to synonyms, paraphrasing, or partially correct yet clinically valid descriptions. To address this limitation, we adopt an \textit{LLM-as-a-Judge} evaluation protocol~\cite{llm_judge,liu2023g}, using GPT-5-mini as the evaluator. Each evaluation instance provides the question, the model prediction, and the reference answer to the judge within a structured rubric, enabling consistent semantic assessment without assuming a fixed lexical space. The judge assigns a score on a five-point scale, where higher scores indicate closer semantic alignment with the reference answer. For frozen (zero-shot) VLMs, we report the mean LLM-based score and LLM-based accuracy, treating scores $\geq 4$ as correct. For fine-tuned models that produce concise single-word outputs aligned with a predefined label vocabulary, we use standard exact-match accuracy and F1-score. Importantly, the same evaluation protocol is applied consistently to both original and less-biased questions, ensuring that observed performance gaps reflect linguistic shortcut reliance rather than evaluation artifacts.

\section{Results and discussions}\label{results}

\subsection{Experimental Setup} 
We evaluate five vision-language models under zero-shot and fine-tuned settings: three general-purpose VLMs, Qwen2.5-VL-7B-Instruct~\cite{bai2025qwen2}, LLaVA-OneVision-1.5-8B~\cite{an2025llava}, and PaliGemma2-10B-Mix~\cite{steiner2024paligemma} and two surgical-specific models PitVQA-Net~\cite{he2024pitvqa} and EndoChat~\cite{endochat}. This setup enables comparison of linguistic shortcut reliance across both general-purpose and domain-specialized architectures. In the zero-shot setting, we evaluate models without additional surgical fine-tuning and report LLM-based Score and Accuracy. In the fine-tuned setting, models are trained on the SurgCheck training split using original QA pairs and evaluated on both original and less-biased test sets, reporting exact-match Accuracy and F1-score to quantify performance degradation when linguistic shortcuts are removed. Surgical-specific models follow their original fine-tuning protocols: PitVQA-Net is trained from scratch, while EndoChat updates only LoRA adapters with frozen vision and language backbones. General-purpose VLMs are fine-tuned for one epoch.

\subsection{Zero-shot Evaluation}

\begin{table}[t]
\centering
\small
\setlength{\tabcolsep}{6pt}
\renewcommand{\arraystretch}{1.2}
\caption{\textbf{Zero-shot performance on SurgCheck benchmark}. Comparison of vision--language models evaluated without surgical fine-tuning. LLM-based Score denotes the mean semantic correctness score (1--5) assigned by the LLM-as-a-Judge. LLM-based Accuracy is computed by treating predictions with scores $\geq 4$ as correct.}
\label{tab:zeroshot}
\begin{tabular}{lcccc}
\toprule
\multirow{2}{*}{Model} & 
\multicolumn{2}{c}{LLM-based Score} & 
\multicolumn{2}{c}{LLM-based Accuracy} \\
\cmidrule(lr){2-3} \cmidrule(lr){4-5}
 & Orig & Less-bias & Orig & Less-bias \\
\midrule
EndoChat & 2.01 & 1.90 {\scriptsize(\textcolor{red}{$\blacktriangledown$0.11})} & 18.24 & 16.40 {\scriptsize(\textcolor{red}{$\blacktriangledown$1.84})} \\
Qwen2.5-VL-7B & 2.22 & 2.10 {\scriptsize(\textcolor{red}{$\blacktriangledown$0.12})} & 28.06 & 24.62 {\scriptsize(\textcolor{red}{$\blacktriangledown$3.44})} \\
LLaVA-OV-1.5-8B & 2.18 & 2.10 {\scriptsize(\textcolor{red}{$\blacktriangledown$0.08})} & 23.08 & 22.72 {\scriptsize(\textcolor{red}{$\blacktriangledown$0.36})} \\
PaliGemma2-10B-Mix & 2.04 & 2.05 {\scriptsize(\textcolor{green}{$\blacktriangle$0.01})} & 20.59 & 21.74 {\scriptsize(\textcolor{green}{$\blacktriangle$1.15})} \\
\bottomrule
\end{tabular}
\end{table}
As shown in Table~\ref{tab:zeroshot}, all models achieve relatively low overall performance, which is expected given their lack of exposure to surgical data. Notably, performance consistently decreases when moving from the original to the less-biased question set, indicating reliance on linguistic shortcuts even in the zero-shot setting. Among general-purpose VLMs, Qwen2.5-VL-7B shows the largest accuracy drop (from 28.06 to 24.62), followed by LLaVA-OV-1.5-8B (23.08 to 22.72). PaliGemma2-10B-Mix exhibits slightly higher robustness, with accuracy increasing marginally from 20.59 to 21.74. The surgical-specific model EndoChat also demonstrates performance degradation (18.24 to 16.40), suggesting that domain-specific pretraining alone does not eliminate sensitivity to linguistic bias. Overall, these results confirm that linguistic shortcuts influence model predictions even before task-specific fine-tuning.

\subsection{Fine-tuned Evaluation}

\begin{table}[]\centering
\caption{\textbf{Fine-tuned performance on the SurgCheck benchmark.}
All VLMs are fine-tuned on the training split using original QA pairs and evaluated on both the original and less-biased test sets.}
  \label{tab:finetune_llm_eval} \small \setlength{\tabcolsep}{6pt} \renewcommand{\arraystretch}{1.2}
  \begin{tabular}{lcccc}
  \hline
  \multirow{2}{*}{Model}   & \multicolumn{2}{c}{Accuracy} & \multicolumn{2}{c}{F1-Score} \\
  \cmidrule(lr){2-3} \cmidrule(lr){4-5}
                            & Orig & Less-bias & Orig & Less-bias  \\ \hline
  PitVQA-Net               & 74.05  & 66.81 {\scriptsize(\textcolor{red}{$\blacktriangledown$7.24})}  & 72.68  & 64.98 {\scriptsize(\textcolor{red}{$\blacktriangledown$7.70})} \\
  EndoChat                 & 62.41  & 55.19 {\scriptsize(\textcolor{red}{$\blacktriangledown$7.22})}  & 59.28  & 51.51 {\scriptsize(\textcolor{red}{$\blacktriangledown$7.77})} \\
  Qwen2.5-VL-7B            & 71.40  & 66.74 {\scriptsize(\textcolor{red}{$\blacktriangledown$4.66})}  & 44.83  & 35.25 {\scriptsize(\textcolor{red}{$\blacktriangledown$9.58})} \\
  LLaVA-OV-1.5-8B          & 71.19  & 63.19 {\scriptsize(\textcolor{red}{$\blacktriangledown$8.00})}  & 46.83  & 35.87 {\scriptsize(\textcolor{red}{$\blacktriangledown$10.96})} \\
  PaliGemma2-10B-Mix       & 77.64  & 73.94 {\scriptsize(\textcolor{red}{$\blacktriangledown$3.70})}  & 56.89  & 51.18 {\scriptsize(\textcolor{red}{$\blacktriangledown$5.71})} \\ \hline
  \end{tabular}
\end{table}
After fine-tuning on the SurgCheck training split, all models show substantial performance improvements compared to their zero-shot results (Table~\ref{tab:finetune_llm_eval}). However, when evaluated on the less-biased question set, where linguistic shortcuts are removed while the visual content and ground-truth answers remain identical, all models experience consistent performance degradation. Among general-purpose VLMs, LLaVA-OV-1.5-8B exhibits the largest accuracy decrease, dropping from 71.19 on the original question set to 63.19 on the less-biased set. Qwen2.5-VL-7B shows a similar trend, with accuracy decreasing from 71.40 to 66.74. PaliGemma2-10B-Mix achieves the highest overall accuracy (77.64) and demonstrates relatively higher robustness, although its performance still declines on the less-biased questions. Importantly, surgical-specific models exhibit comparable degradation patterns. PitVQA-Net drops from 74.05 to 66.81, and EndoChat from 62.41 to 55.19 when linguistic shortcuts are removed. These results indicate that while fine-tuning improves task adaptation, neither domain-specific architectures nor specialized training strategies eliminate reliance on linguistic shortcuts in surgical VQA.

\subsection{Per-Category Analysis}
\begin{figure}[htbp]
\centering
\includegraphics[width=\textwidth]{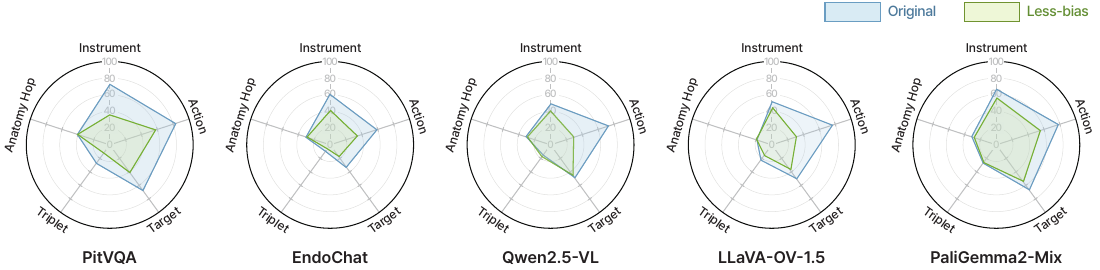}
\caption{\textbf{Category-wise F1-score across five VLMs.} Radar plots compare performance on original (blue) and less-biased (green) questions for task categories where linguistic bias is controlled. Larger gaps indicate stronger
linguistic shortcut reliance.}
\label{fig:cat_f1}
\end{figure}

We analyze category-wise performance to identify which types of surgical questions are most affected by linguistic shortcuts (Fig.~\ref{fig:cat_f1}). This analysis focuses on task categories with bias-controlled paired questions: Instrument, Action, Target, Triplet, and Anatomy-hop. Categories such as Instrument Count, Anatomy, and CVS are excluded, as original and less-biased questions are identical. Across all models, Action and Target tasks show the largest F1-score drops when linguistic cues are removed, indicating strong reliance on entity names for predicting action–target associations. This pattern is consistent across both general-purpose and surgical-specific models. In contrast, Anatomy-hop questions exhibit minimal degradation, as they require spatial reasoning over anatomical relationships that cannot be resolved from linguistic cues alone. Instrument and Triplet tasks show intermediate behavior, with moderate but consistent performance drops. Overall, these results indicate that linguistic shortcuts primarily affect action- and target-centric questions, whereas tasks requiring spatial reasoning are comparatively robust.
\subsection{Per-Cue-Type Analysis}
\begin{figure}[t]
\centering
\includegraphics[width=\textwidth]{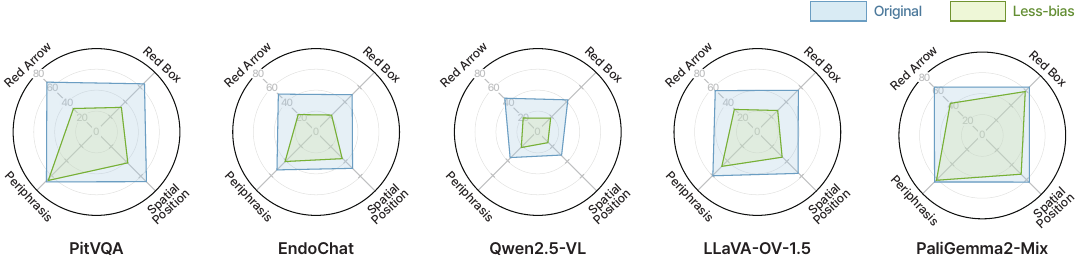}
\caption{\textbf{Cue-type-wise F1 performance across five VLMs.}
Radar plots compare F1-scores on original (blue) and less-biased (green)
questions for each grounding cue type. Larger gaps indicate stronger
linguistic shortcut reliance.}
\label{fig:cue_type_f1}
\end{figure}

We analyze how different grounding cue types influence model performance (Fig.~\ref{fig:cue_type_f1}). To ensure a fair comparison, we construct a balanced evaluation subset containing only frames for which all four cue types are applicable. For each frame, we generate four versions of the less-biased question that differ only in cue type, while keeping the visual content, question structure, task category, and ground-truth answer identical. Consequently, cue type is the only varying factor in this analysis. We report F1-score to account for class imbalance across task categories and to better reflect performance changes under bias-controlled evaluation. Across all models, explicit grounding cues (arrow and bounding box) exhibit the largest F1-score drops relative to original questions. Notably, this degradation occurs despite precise visual localization, revealing deficiencies in visual grounding rather than ambiguity in reference. Even when the target region is explicitly indicated, models struggle to associate localized visual evidence with the correct action or target in the absence of entity names. In contrast, periphrasis cues show the smallest performance degradation. By referring to instruments through contextual relationships (e.g., the operator’s hand), these cues encourage models to integrate visual context rather than rely solely on lexical shortcuts. Spatial position cues exhibit intermediate behavior, requiring coarse spatial reasoning within the image frame while still allowing partial shortcut exploitation. Overall, these results demonstrate that cue explicitness plays a critical role in diagnosing shortcut-driven behavior in surgical VQA. Importantly, reliance on entity names persists even under strong visual localization, highlighting a fundamental limitation in current VLMs’ ability to ground predictions in visual evidence alone.
\subsection{Text-only Ablation}
\begin{figure}[t]
\centering
\includegraphics[width=\textwidth]{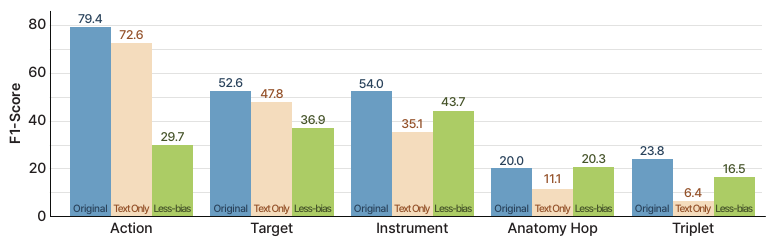}
\caption{\textbf{Text-only ablation with category-wise F1-scores under three conditions (Original, Text-only, Less-bias).} Performance of fine-tuned LLaVA-OV-1.5-8B on original questions with image (blue), original questions without image (text-only, orange), and less-biased questions with image (green).}
\label{fig:text_only}
\end{figure}

To directly test whether high performance on original questions reflects genuine visual reasoning, we conduct a text-only ablation using the fine-tuned LLaVA-OV-1.5-8B model (Fig.~\ref{fig:text_only}). In this setting, images are removed at inference time while the original question text is preserved, and performance is compared against original and less-biased conditions. For Action and Target tasks, text-only performance remains close to that of the original setting, with Action accuracy decreasing only from 79.4 to 72.6 and Target accuracy from 52.6 to 47.8, despite the absence of visual input. Performance drops sharply on the less-biased questions, falling to 29.7 for Action and 36.9 for Target, even though the image is retained. This indicates that, in Action and Target tasks, strong performance under the original setting is largely driven by linguistic shortcuts rather than visual evidence.
By comparison, Anatomy-hop and Triplet tasks exhibit the opposite pattern: performance on less-biased questions remains relatively stable compared to the original setting and exceeds that of the text-only condition, reaching 20.3 and 16.5, respectively, compared to 11.1 and 6.4 under text-only evaluation. These tasks require spatial or compositional reasoning that cannot be resolved from text alone, thereby limiting linguistic shortcuts and forcing reliance on visual input. Overall, this ablation provides direct causal evidence that linguistic shortcuts enable models to bypass visual reasoning for action- and target-centric questions, while tasks requiring genuine spatial understanding remain dependent on visual evidence.

\subsection{Qualitative Analysis}

\begin{figure}[t]
\centering
\includegraphics[width=\textwidth]{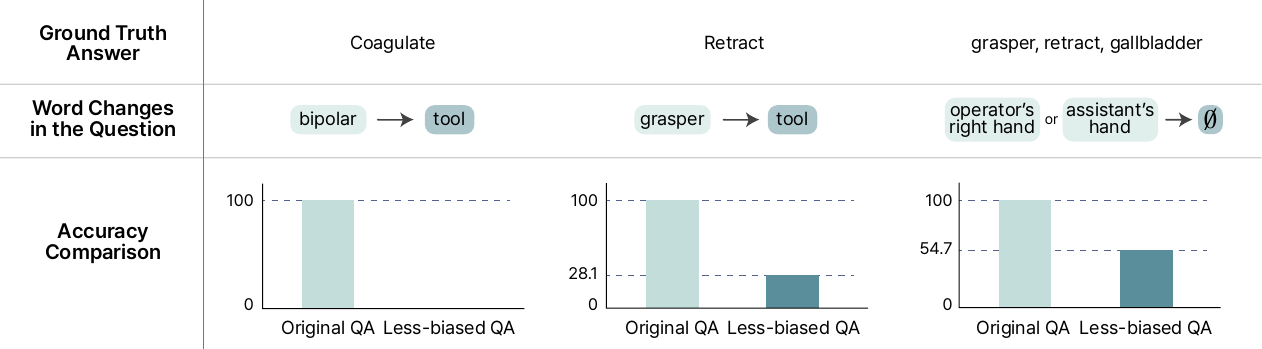}
\caption{\textbf{Top-3 performance gap between original and less-biased QAs for fine-tuned LLaVA-OV-1.5-8B}.
Small wording changes from the original to the less-biased
questions (e.g., replacing explicit instrument names with “tool” or removing
hand references) lead to substantial accuracy drops, despite identical visual
content and ground-truth answers.}
\label{fig:llava_top3}
\end{figure}
Fig.~\ref{fig:llava_top3} shows representative failure cases of the fine-tuned LLaVA-OV-1.5-8B model. In all examples, the image and ground-truth answer remain unchanged, while only a small linguistic modification is applied to the question. Replacing explicit instrument names (e.g., “bipolar”, “grasper”) with generic terms (“tool”) consistently causes incorrect predictions, indicating strong reliance on lexical associations rather than visual grounding. Similarly, removing referential cues such as the operator’s right hand or the assistant’s hand often leads to incorrect instrument–action or instrument–target associations. These examples qualitatively confirm that model failures arise from the removal of linguistic shortcuts, not from visual ambiguity, reinforcing our quantitative findings.
\section{Conclusion} 
We presented \textbf{SurgCheck}, a diagnostic benchmark for surgical visual question answering (VQA). SurgCheck employs paired-question evaluation and grounding cues to systematically disentangle linguistic shortcut reliance from visual reasoning. Built upon structured scene-graph annotations, it enables standardized assessment across eight surgical understanding tasks spanning observation and reasoning levels. Experiments on both general-purpose and surgical-specific VLMs show that strong performance on conventional surgical questions does not necessarily indicate genuine visual understanding. Across both zero-shot and fine-tuned settings, models consistently degrade on less-biased questions despite identical visual inputs and ground-truth answers. Per-category analysis shows that questions centered on actions and targets are especially susceptible to linguistic shortcuts, while Anatomy-hop and Triplet questions remain comparatively robust. By combining controlled bias cues with a reproducible LLM-as-a-Judge evaluation protocol, SurgCheck establishes a consistent and interpretable framework for diagnosing visual reasoning in surgical VQA.

We hope SurgCheck will support more reliable evaluation of multimodal models and encourage progress toward methods that genuinely ground predictions in visual evidence.

\subsubsection*{Acknowledgements.}
This work was supported in part by the Institute of Information and Communications Technology Planning and Evaluation (IITP) grant funded by the Korea government (MSIT) under Grant IITP-2025-RS-2023-00258649 (ITRC support program), by the National Research Foundation of Korea (NRF) grant funded by the Korea government (MSIT) (No. RS-2024-00334321, RS-2024-00392495), and by the ‘Future Medicine 2030’ project of Samsung Medical Center (SMX1230771).

\bibliographystyle{splncs04}
\bibliography{resource}

\end{document}